\documentclass[preprint,12pt]{elsarticle}

\usepackage{graphicx}

\newcommand{\NEWTRIPLET}{LTO}
\newcommand{\NEWPAIR}{LPO}
\newcommand{\NEWDRUG}{LDO}
\newcommand{\NEWCELL}{LCO}

\usepackage{amssymb}
\begin{document}

\begin{frontmatter}



\title{New methods for drug synergy prediction: a mini-review}


\author[inst1]{Fatemeh Abbasi}

\affiliation[inst1]{organization={Laboratory of Bioinformatics and Drug Design (LBD),  Institute of Biochemistry and Biophysics},%Department and Organization
            addressline={University of Tehran}, 
            city={Tehran},
            %postcode={00000}, 
            %state={State One},
            country={Iran}
            }
\ead{abbasi.f@ut.ac.ir}
\author[inst2]{Juho Rousu}
%\author[inst1,inst2]{Author Three}

\affiliation[inst2]{organization={Department of Computer Science},%Department and Organization
            addressline={Aalto University}, 
            city={Espoo},
%            postcode={00076}, 
%            state={State Two},
            country={Finland}}
\ead{juho.rousu@aalto.fi}

\begin{abstract}
In this mini-review, we explore the new prediction methods for drug combination synergy relying on high-throughput combinatorial screens. The fast progress of the field is witnessed in the more than thirty original machine learning methods published since 2021, a clear majority of them based on deep learning techniques. We aim to put these papers under a unifying lens by highlighting the core technologies, the data sources, the input data types and synergy scores used in the methods, as well as the prediction scenarios and evaluation protocols that the papers deal with. Our finding is that the best methods accurately solve the synergy prediction scenarios involving known drugs or cell lines while the scenarios involving new drugs or cell lines still fall short of an accurate prediction level.
\end{abstract}



%\begin{keyword}
%keyword one \sep keyword two
%\end{keyword}

\end{frontmatter}


\section{Introduction}
\label{sec:sample1}

Combination therapies involving two or more drugs are nowadays frequently used to treat complex diseases. Combination therapies can enhance treatment efficacy while mitigating side-effects and drug resistance, due to lower required drug doses and drug synergistic effects. The vast number of potential drug combinations presents a major bottleneck for developing new combination therapies, which calls for new computational approaches to facilitate the exploration of drug combination spaces. 

In recent years, several large high-throughput screening datasets for drug combination 
have been published \cite{merck2016,almanac2017,az-sanger2019,zheng2021drugcomb,liu2020drugcombdb} which has fueled the development of a new generation of predictive models for drug synergy.

In this mini-review we focus on the new prediction methods based on deep learning, pioneered by the DeepSynergy method \cite{preuer2018deepsynergy}, and  machine learning developed in last years. We put these papers under a unifying lens by first highlighting the prediction setups, included the prediction scenarios, the data sources, the input data types and synergy scores used in model output, as well as the evaluation protocols used. Following that, we discuss the core technologies, commonly used in the models, and present an overview of the predictive performance of the models. 

 We limit our focus to new methods that have been tested against large benchmark datasets on pre-clinical synergy and dose response data, in particular leaving out drug-drug interaction (DDI) prediction, which has its own focused literature (see e.g. \cite{vo2022road}), as well as papers that are using small scale data or lacking comparison to alternative prediction methods. In addition, we do not cover web-based tools, software and libraries that implement prediction methods (see e.g. \cite{wu2022machine} for a review).

\section{Prediction Setups}

\paragraph{Synergy scores} The output of a synergy prediction model is either a real-valued synergy score (a regression task) or a binary prediction (synergistic/non-synergistic), which is generally obtained from the synergy scores by thresholding. The most common synergy scores are Bliss independence, Loewe additivity, zero interaction potential (ZIP), and Highest single agent (HSA), ComboScore and S-score \cite{malyutina2019drug}. Synergy prediction models can be trained separately against different synergy scores or, by using multi-task learning \cite{el2023marsy}, trained against multiple scores at the same time. Regression methods generally use the synergy scores as the model output, while classification methods use some form of thresholding to classify the combinations as synergistic, non-synergistic or antagonistic. Alternatively, the methods can be trained to predict the dose-response of the combination, and the synergy scores can be computed in a post-processing step \cite{julkunen2020leveraging}.

\begin{table}[htbp]
\caption{Data sources containing combination response data for model training and evaluation (top part) and sources containing additional input data for drugs and cell lines (bottom part). The data sources marked with '*' are databases integrating multiple studies.}
\label{table:datasources}
\small
\begin{tabular}{|p{2.5cm}p{4.8cm}p{4.5cm}c|}
\hline
\textbf{Data source} & \textbf{Type of Data} & \textbf{Size}  & \textbf{Ref} \\
\hline

\hline
AZ-SANGER  & 6-by-6 dose-response matrix  & 118 drugs, 85 cell lines, 910 pairs, 11576 triplets   & \cite{az-sanger2019} \\
GDSC$^2$ & 2-by-7 dose-response matrix  & 65 drugs, 125 cell lines, 2025 pairs, 108259 triplets & \cite{jaaks2022effective} \\
DrugComb$^*$ & synergy scores & 8397 drugs, 2320 cell lines, 739964 triplets &  \cite{zheng2021drugcomb} \\
DrugCombDB$^*$ & synergy scores &  2887 drugs, 124 cell lines, 448 555 triplets & \cite{liu2020drugcombdb} \\
Merck2016  & 4-by-4 dose-response matrix & 39 drugs, 38 cancer cell lines, 583 pairs, 22737 triplets & \cite{merck2016} \\
NSCLC & 1-by-5 dose-response & 263 drugs, 81 cell lines, 5082 pairs, 369776 triplets & \cite{nair2023landscape} \\
NCI-ALMANAC & 3-by-3 or 5-by-3 dose-response matrix &  104 drugs, 5232 pairs, 60 cell lines, 304549 triplets & \cite{almanac2017} \\
%SynergyFinder & Synergy scores & \textgreater{}6,000 drug combinations & * \\
SynergXDB$^*$ & dose-response matrix, cell line profile & 1977 drugs, 151 cell lines, 22507 pairs, 477839 triplets & \cite{seo2020synergxdb} \\
CTD2-NCATS & %Efficacy 
synergy scores & \textgreater{}11000 drug combinations & \cite{aksoy2017ctd2} \\
\hline
CCLE & CLE data %for individual drugs and drug combinations 
 on the expression levels of genes in different cell lines & \textgreater{}1,000 cell lines & \cite{ghandi2019next} \\

CTRP & Multi-omics data of cancer cell lines, dose response of individual drugs & 481 compounds, 860 cancer cell lines & \cite{rees2016correlating} \\
GDSC & Genomic, transcriptomic, and drug sensitivity data & \textgreater{}1,000 cancer cell lines & \cite{yang2012genomics} \\
DepMap & Gene dependency data from knockout experiments & \textgreater{}1,500 cell lines & \cite{DepMap2023} \\

Drugbank & molecular information about drugs, their mechanisms, interactions, and targets & \textgreater{}500,000 drugs and drug products &\cite{wishart2018drugbank} \\

PubChem & chemicals by name, molecular formula and structure  & \textgreater{}166M compounds &\cite{kim2023pubchem} \\
\hline
\end{tabular}
\end{table}

\paragraph{Input data}

Drug combination prediction models utilize diverse data types to capture the complexity of drug interactions and their effects. Table \ref{table:datasources} lists the most important data sources for drug synergy prediction, including  synergy or dose-response data of drug combinations (top part) and data sources containing descriptors and profiling data of drugs and cell lines (bottom part).

Below we explain shortly the most common data types used by the methods included in this mini-review, to illustrate the commonalities and differences of the predictive methods listed in Table \ref{table:methods}. We note that there is a large amount of further detail and variations within the general datatypes and details should be checked from the original papers.

\begin{itemize}
\item Drug features: These encompass various molecular descriptors and fingerprints (FP), chemical structures (CS), pharmacological properties, drug (monotherapy) dose-response. Such features provide information about the characteristics and potential interactions of the drugs. In addition, some methods include drug combination response in other cell lines than the one being predicted as an input feature \cite{julkunen2020leveraging, wang2021modeling,ronneberg2023dose}.

\item Genomic and transcriptomic data: Cell line  expression profiles (CLE), miRNA expression, genomic mutations (MUT), copy number variations (CNV), and other genomic data can be leveraged to identify molecular signatures associated with drug response. These multi-omics data sources contribute to a deeper understanding of the mechanisms underlying drug combination effects.

\item Biological pathways and networks: Protein-protein interaction networks (PPI) and drug-target associations (DTA), offer valuable insights into the underlying mechanisms of drug combinations. They enable the incorporation of biological knowledge and context into the prediction models. 

\end{itemize}

The methods in Table \ref{table:methods} generally divide into two types: (1) narrow input data, relying on one type of drug feature and one type of cell line feature (usually CLE) combined with very large training sets, and (2) broad input data, using multiple drug, genomic and biomedical pathway data, but with smaller training data sets. The first category generally allows easier scale-up to large data better but may be restricted in generalizing in the more challenging prediction scenarios (\NEWDRUG,\NEWCELL) due to limited biological context. The second category, on the other hand, has broader biological context but may be restricted in training data size.

\paragraph{Prediction scenarios.} The difficulty of synergy prediction depends significantly on the assumption of which data we expect to be present at prediction time. Given a triplet 
$(D_1,D_2,C)$ of a pair of drugs ($D_1$,$D_2$) and a cell line ($C$) to be predicted in the test set, the following scenarios are frequently studied:
\begin{itemize}
 \item Leave-Triplet-Out (\NEWTRIPLET): The triplet $(D_1,D_2,C)$ does not occur in the training data. However, the drug pairs ($D_1$,$D_2$) may occur in the training set connected to another cell line $C'$.
 \item Leave-Pair-Out (\NEWPAIR): The drug pair ($D_1$,$D_2$) does not occur in the training set in connection to any cell line. however, the individual drugs may occur in training set in connection to any cell line.
 \item Leave-Drug-Out (\NEWDRUG): At least one of the drugs in the pair $(D_1,D_2)$ does not occur in training set at all.
 \item Leave-Cell line-Out (\NEWCELL): The cell line $C$ does not occur in the training set but drugs $D_1$ and $D_2$ may occur in the training set in conjunction of other cell lines.
\end{itemize}

The choice of the scenario has a major effect on the accuracy of the prediction. However, in the papers describing the methods (Table \ref{table:methods}), the assumed scenario is often only implicitly given, in the description of the training-validation-test splitting or the cross-validation procedure. Table \ref{tab:validation-methods} lists the scenarios that we identified in the papers. By far the most common scenario is \NEWTRIPLET\ which follows from splitting unique triplets randomly into training, validation and test sets or cross-validation folds, followed by 
\NEWPAIR\ \cite{zhang2021synergistic, hosseini2023ccsynergy, xu2023dffndds, hu2022dtsyn, el2023marsy, zhang2023mgae, PRODeepSyn, li2023snrmpacdc, preto2022synpred}, while \NEWDRUG\  \cite{lin2022enhanced, zhang2023mgae, li2023snrmpacdc, preto2022synpred, liu2021transynergy} and \NEWCELL\ \cite{lin2022enhanced, zhang2023mgae, pinoli2021predicting, preto2022synpred, liu2021transynergy} have received less attention, perhaps partly due to the that each scenario requires its dedicated data splitting strategy and partly due to their difficulty. 

\paragraph{Evaluation protocols and hyperparameter tuning}

The most popular cross-validation procedure among the references is 5-fold cross-validation, where the splits are chosen to honor the scenario under investigation. Another common strategy is random train-test-split, which is repeated a few times. A few papers use in addition an independent test set that is not used in the model development \cite{wang2022deepdds, rafiei2023deeptrasynergy, liu2022hypergraphsynergy, kim2021anticancer, brahim2021matchmaker, PRODeepSyn}. 

 Most synergy prediction models have some hyperparameters that are given as input to the learning algorithm, which typically affects the capacity to the model to fit the data. However, over half of the papers in this review did not clearly explain the data splitting strategy for hyperparameter tuning (Table \ref{tab:validation-methods}). The use of a validation set separated from the training and test data is a sound approach used in several of the papers. A few papers \cite{wang2023deml, brahim2021matchmaker, PRODeepSyn, liu2021transynergy} employ a rigorous nested cross-validation approach, where the inner loop is used to tune the hyperparameters and the outer one is used for obtaining the performance estimates.  

 %\end{document}

\begin{table}[p]
\tiny
\centering
\caption{Summary of Recent Methods in Drug Synergy Prediction}
\label{table:methods}
%\begin{tabular}{|p{2cm}|p{0.5cm}|p{2.1cm}|p{2.5cm}|p{2.5cm}|p{1.5cm}|p{2cm}|p{0.7cm}|}
\begin{tabular}{|p{1.7cm}lp{1.7cm}p{3cm}p{2cm}p{1.7cm}c|}
\hline
\textbf{Method} & \textbf{Year} & \textbf{Model type} & \textbf{Data Sources} & \textbf{Input Datatypes} & \textbf{Synergy scores} &  \textbf{Ref} \\ \hline
%AI-DrugNet& 2023 & GCN & DrugCombDB & SMILES, CLE, GRN & Classification &\NEWTRIPLET & \cite{pan2023ai}\\ %\hline 
AuDNNSynergy & 2021 & AE,DNN & Merck-2016, TCGA  & FP, MUT, CNV & Loewe &  \cite{SynergisticDeepLearningModels}\\ %\hline
CCSynergy & 2023 & DNN & Merck-2016, DrugComb, Chemical Checker & FP, multi-omics & Loewe & \cite{hosseini2023ccsynergy} \\ %\hline
comboLTR & 2021 & HOFM & NCI-ALMANAC & MACCS, multi-omics & Dose-response  & \cite{wang2021modeling}\\ %\hline
DCE-DForest & 2022 & BERT,Deep Forest & NCI-ALMANAC & SMILES, CLE & Classification &  \cite{zhang2022dce}\\ %\hline
DeepDDS & 2022 & GNN, DNN & Merck-2016 & CS, CLE & Loewe & \cite{wang2022deepdds} \\ %\hline
DeepTraSynergy & 2023 & Multitask, Transformer & DrugCombDB, OncologyScreen & PPI, SMILES & Loewe  & \cite{rafiei2023deeptrasynergy}\\
DEML & 2023 & DNN, Ensemble & DrugComb & FP, CLE & Loewe &  \cite{wang2023deml}\\ %\hline
DFFNDDS & 2023 & BERT, DNN & DrugComb, DrugCombDB & SMILES, FP, CLE & Loewe  & \cite{xu2023dffndds}\\ %\hline

DTSyn & 2022 & Transformer & Merck-2016, AstraZeneca, FLOBAK, ALMANA, FORCINA, YOHE , CCLE, DeepChem& CLE, SMILES, PPI & Loewe  & \cite{hu2022dtsyn} \\ %\hline
EC-DFR & 2022 & Deep Forest &DrugComb, LINCS, PubChem &CLE, FP & S score  &\cite{lin2022enhanced} \\ %\hline
Forsyn & 2023 & Deep Forest & NCI-ALMANAC, DrugComb, DrugCombDB, AZ-Sanger & SMILES, CLE & Classification  & \cite{wu2023hybrid}\\ %\hline
GraphSynergy & 2021 & GCN & DrugCombDB, OncologyScreen, CCLE & PPI, DPA,CPA & Loewe  & \cite{yang2021graphsynergy}\\ %\hline
GAECDS & 2023 & Graph AE, CNN & DrugComb, CCLE, PubChem & CLE, SMILES & Loewe  & \cite{li2023predicting} \\
Hypergraph-Synergy & 2022 & GNN & Merck-2016, NCI-ALMANAC, PubChem, COSMIC & SMILES, CLE &Loewe & \cite{liu2022hypergraphsynergy} \\ %\hline
HyperSynergy & 2023 & GNN, DNN, CNN & SYNERGxDB, CCLE, PubChem & SMILES, CLE & Loewe &  \cite{zhang2023few} \\ %\hline
KGE-DC & 2022 & GNN & DrugComb, DrugBank & FP, CLE, DPA & Loewe, Classification  &\cite{zhang2022knowledge} \\
Kim et al. & 2021 & DNN, Transfer learning & DrugComb, CCLE & FP, CLE &Loewe &  \cite{kim2021anticancer}\\ %\hline
Ma et al. & 2021 & PCA,DNN & AZ-SANGER, Merck2016 & FP, CLE &  Loewe  & \cite{ma2021prediction}\\ %\hline
MARSY & 2023 & DNN, Multitask & DrugComb & CLE & ZIP, S-score &\cite{el2023marsy} \\ %\hline
MatchMaker & 2022 & DNN & DrugComb & CS, CLE & Loewe &\cite{brahim2021matchmaker} \\ %\hline
MGAE-DC & 2023 & GNN,AE, DNN & Merck-2016, NCI-ALMANAC, CLOUD, FORCINA & FP & Loewe,
Bliss,
ZIP,
HSA &  \cite{zhang2023mgae} \\ %\hline
Nafshi et al. & 2021 & PMF & NCI-ALMANAC & Dose-response & Dose-response &  \cite{nafshi2021predicting}\\ %\hline
NEXGB & 2022 & XGboost & DrugCombDB & DPA, PPI, Pathway & Classification  & \cite{meng2022nexgb} \\ %\hline
PIICM & 2023 & GP & Merck-2016 & Dose-response & Dose-response & \cite{ronneberg2023dose} \\
Pinoli et al. & 2022 & NMTF & DrugCombDB & CLE, CNV & ZIP &\cite{pinoli2021predicting}\\ %\hline
PRODeepSyn & 2022 & GCN & Merck-2016  & FP, PPI, CLE, MUT & Loewe &\cite{PRODeepSyn} \\ %\hline
SDCNet & 2022 & GCN & DrugComb & FP  & Loewe  &\cite{zhang2022predicting} \\ %\hline
Shim et al. & 2022 & Word2vec, DNN & NCI-ALMANAC & Pubmed & ComboScore & \cite{shim2022novel}\\ %\hline
SNRMPACDC & 2023 & Siamese Network & Merck-2016 & FP, CNV, MUT  & Loewe  &\cite{li2023snrmpacdc}\\ %\hline
SynPathy & 2022 & DNN & DrugComb & Pathway, CS & Loewe & \cite{tang2022synpathy} \\ %\hline
SYNPRED & 2022 & Ensemble & NCI-ALMANAC, DrugComb & FP, multi-omics & Loewe  & \cite{preto2022synpred}\\ %\hline
SynPredict & 2023 & DNN & Merck-2016, NCI-ALMANAC & FP, CLE & Loewe, Bliss, ZIP, HSA, S-score   &\cite{alsherbiny2021trustworthy}\\ %\hline
TranSynergy & 2021 & DNN, Transformer & Merck-2016, Drugbank, Chembel, CCLE, GDSC DepMap & CLE, SMILES & Loewe  & \cite{liu2021transynergy} \\ 
\hline
\end{tabular}
\end{table}

\begin{table}[hbt!]
 \centering
 \caption{Summary of evaluation methods in the reviewed papers, including the prediction scenarios, validation protocols and hyperparameter tuning.}
\label{tab:validation-methods}
 \footnotesize
\begin{tabular}{|p{3cm}lp{4.5cm}p{3cm}|}
 \hline
 \textbf{Method} & \textbf{Scenarios} & \textbf{Validation Method} & \textbf{Hyperparameter Tuning}  \\
 \hline
AuDNNSynergy & LPO & 5-fold CV & validation set  \\ 
CCSynergy & LPO & 5-fold CV &  not clear\\
comboLTR & LPO & 5-fold CV & validation set   \\
DCE-DForest & LTO & repeated train-test & validation set \\
DeepDDS & LTO & 5-fold CV, independent set & not clear   \\
DeepTraSynergy & LTO & 5-fold CV, independent set & not clear   \\
DEML & LTO& 5-fold CV & nested CV  \\
DFFNDDS  & LPO & 5-fold CV & not clear   \\
DTSyn & LPO & 5-fold CV & not clear \\
EC-DFR & LDO, LCO & 5-fold CV & not clear \\
Forsyn & LTO & 5-fold CV & not clear \\
GraphSynergy  & LTO & not clear & not clear   \\
GAECDS  & LTO & 5-fold CV & not clear  \\
HyperGraphSynergy & LCO, LPO, LTO & 5-fold CV, independent set & not clear  \\
HyperSynergy & LTO & repeated train-test & validation set \\
KGE-DC & LTO & 10-fold CV & not clear  \\
Kim et al. & LTO & 5-fold CV, independent set & not clear   \\
Ma et al. & LTO & not clear & not clear   \\
MARSY & LPO, LTO & 5-fold CV & validation set  \\
MatchMaker & LPO & 5-fold CV, independent set & nested CV   \\
MGAE-DC & LCO, LDO, LPO, LTO & 10-fold CV & validation set  \\
Nasfi et al. & Custom & train-test split & not clear  \\
NEXGB & LTO & 5-fold CV & not clear \\
PIICM & LTO & train-test split & 5-fold CV  \\
Pinoli et al. & LCO & not clear & not clear \\
PRODeepSyn & LPO & 5-fold CV, independent set & nested CV    \\
SDCNet & LTO & 10-fold CV & not clear  \\
Shim et al. & LTO & 5-fold CV & not clear \\
SNRMPACDC & LPO LDO & 5-fold CV & not clear \\
SynPathy  & LTO & 10-fold CV & validation set  \\
SYNPRED & LCO, LDO, LPO, LTO & train-test split & 3-fold CV \\
SynPredict & LTO & 5-fold CV & not clear \\
TranSynergy & LCO, LDO, LPO & 5-fold CV & nested CV  \\
\hline
\end{tabular}
\end{table}

\section{Predictive models for drug synergy}

Here we provide a short overview of the modeling approaches used in the reviewed papers. The interested reader may check the original papers for details.

\subsection{Neural Network Models}

The majority of the recent drug synergy prediction models (Table \ref{table:methods}) are based on neural networks of various types, which have become popular in recent years. Their main benefit is the ability to learn new representations from large training data while the main deficiency is the computational complexity of training the models.

\textit{Deep Neural Networks (DNN)} are a widely used neural architecture in drug synergy prediction models, either as a component or a standalone model \cite{zhang2021synergistic,hosseini2023ccsynergy,wang2022deepdds, xu2023dffndds, zhang2023few, kim2021anticancer, ma2021prediction, el2023marsy, brahim2021matchmaker, zhang2023mgae, shim2022novel, tang2022synpathy,preto2022synpred}. DNNs are composed of multiple layers of interconnected computation units that compute a linear transform of the inputs followed by a non-linear activation function. Myriads of DNN architectures can be created by using different types of connection patterns between the units. 

\textit{Encoder-decoder networks} such as auto-encoders (AE) and transformers are used to learn latent representations of structured data such as SMILES sequences  or molecular graphs. \textit{Transformers} learn mappings between general inputs and outputs. By relying on attention units, transformers can adaptively focus on different parts of a structured object. In drug synergy prediction, BERT transformer has been used to learn latent representations from SMILES strings \cite{zhang2022dce,xu2023dffndds}. More generally, transformers have been used to map embeddings of input data sources into intermediate representations \cite{liu2021transynergy,rafiei2023deeptrasynergy,hu2022dtsyn}. In addition, the Word2vec encoder-decoder network has been used to extract latent representations from text documents describing drug combinations \cite{shim2022novel}.

\textit{Graph Neural Networks (GNN)} specialize in analyzing  relational data represented as graphs or networks. GNNs are capable of learning embeddings for individual nodes and edges as well as complete graphs. The main benefit of GNNs over text (e.g. SMILES) or vectorial representations (e.g. molecular fingerprints), is their capability to learn fine-grained representations that are still explainable in graphical form. In drug synergy prediction, GNNs are used to model  molecular graphs as well as biological networks of drugs, targets and cell lines \cite{wang2022deepdds,liu2022hypergraphsynergy,ren2022multidrug}. \textit{Graph Convolutional Network (GCN)} is one of the most widely used type of GNN \cite{zhang2022predicting, chen2022drug, PRODeepSyn, yang2021graphsynergy, xu2023dffndds,jiang2020deep}. Graph Attention Networks (GAT) combine graph convolution with attention units for added flexibility \cite{wang2022deepdds}.

\textit{Siamese Network} share parameters between subnetworks processing different data items arising from paired objects, such as pairs of drugs \cite{li2023snrmpacdc}. They are particularly valuable for assessing the similarity or complementarity of drug properties, an essential factor in predicting drug synergy.
 
\subsection{Forest-based models}

Models based on ensembles of trees such as random forest, based on bootstrap aggregation, and XGBoost \cite{meng2022nexgb} based on gradient boosting are strong predictors and frequently used in drug combination prediction. These methods have been recently extended to deep forests where several layers of forests are used \cite{wu2023hybrid, zhang2022dce, lin2022enhanced} for synergy classification. However, similarly to neural networks, they can be computationally intensive to build and challenging to interpret.

\subsection{Factorization models}

Factorization approaches in drug combination prediction involve the decomposition of multi-dimensional tensors into latent factors to extract latent features and relationships between drugs \cite{pinoli2021predicting,nafshi2021predicting}. 
These models are powerful in predicting missing values in incomplete data tensors (e.g. combination response or synergy data) by learning from the co-occurrences of subsets of variables. In particular Higher-Order Factorization Machines (HOFM) \cite{julkunen2020leveraging} and latent tensor reconstruction \cite{wang2021modeling} are accurate in the \NEWTRIPLET\ scenario as well in completion of individual dose-response matrices. On the other hand, they are not expected to have an advantage in the \NEWDRUG\ and \NEWCELL\ scenarios, which require extrapolation outside the known drug or cell line space. 

\subsection{Bayesian models}
Bayesian models provide a consistent fully probabilistic inference approach for modeling drug combination experiments and predicting dose-response relationships. In particular, Gaussian processes have been used to model drug synergy \cite{bayesynergy,ronneberg2023dose}. These models particularly excel in allowing the prediction uncertainty to be rigorously addressed.
However, the complexity of Bayesian inference can be computationally demanding and may require specialized knowledge in statistics, limiting their accessibility and use in broader applications.

\begin{figure}
    \centering
    \includegraphics[width=1\textwidth]{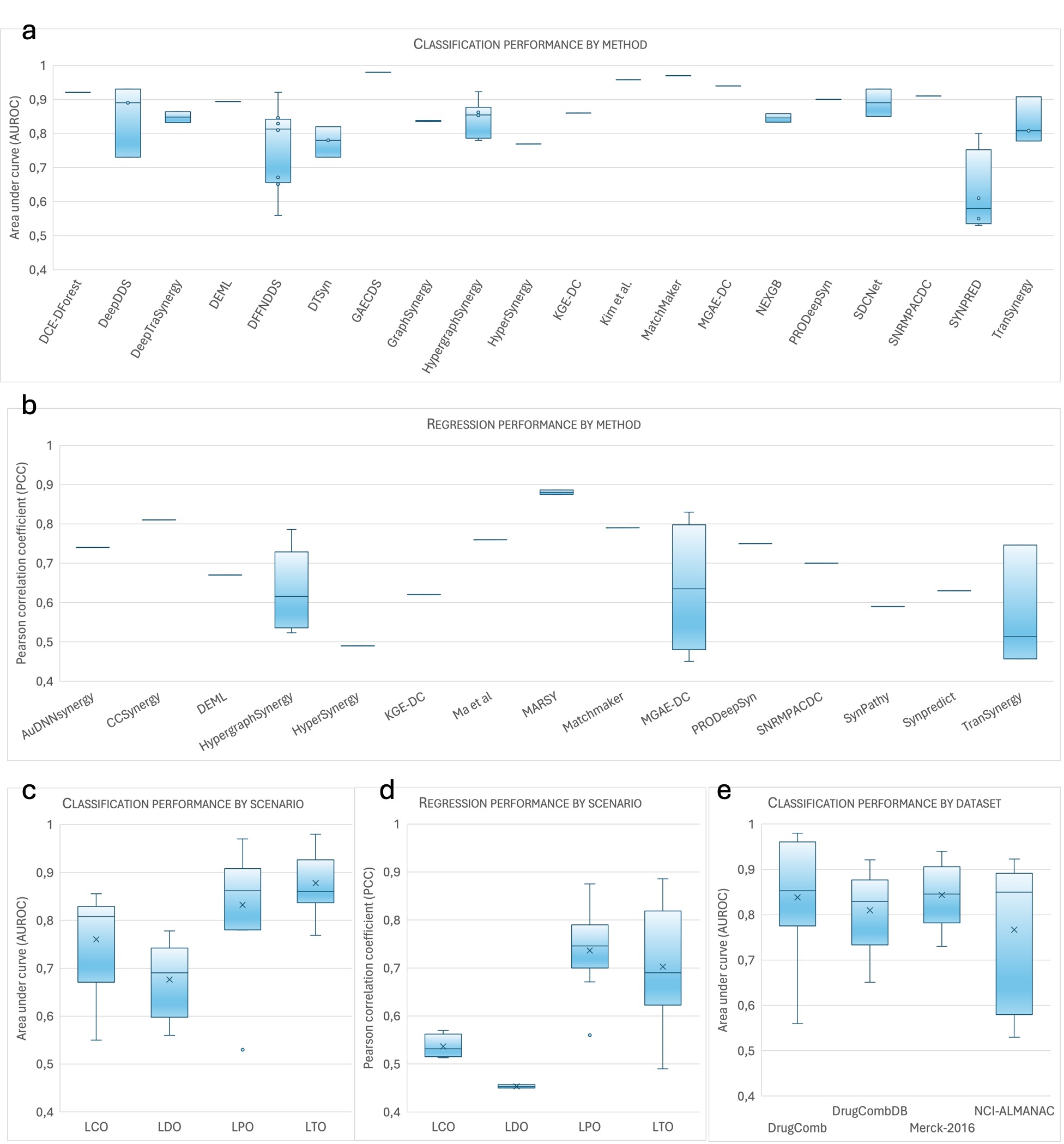}
    \caption{Summary of predictive performance of synergy prediction models: (a) Classification performance by method, (b) Regression performance by method (c) Classification performance by scenario, (d) Regression performance by scenario (e) Classification performance by dataset.}
    \label{fig:perf}
\end{figure}

\section{Prediction performance of the models}

Figure \ref{fig:perf} summarizes the predictive performance of the models reviewed here. Figure \ref{fig:perf}a depicts the reported AUROC values for each classification method. The highest values are due to GAECDS (0.98),  MatchMaker (0.97) and Kim et al. (0.96) while nine further methods reach 0.9 AUROC or more. 
Methods that report performance over multiple scenarios exhibit a range of values given by the Box-Whisker plot. Among regression models predicting Loewe synergy (Figure\ref{fig:perf}b), MARSY obtains the highest Pearson Correlation Coefficient (PCC) of 0.89 on DrugComb, above the second highest value 0.83 by MGAE-DC on Merck-2016, both of these obtained in the LTO scenario. Figures  \ref{fig:perf}c and  \ref{fig:perf}d depict the distributions of AUROC and PCC values in different scenarios. The LCO and LDO scenarios show as significantly harder prediction tasks than LPO and LTO: based on Wilcoxon rank-sum test with Bonferroni correction, LDO has significantly lower AUROC compared to LPO (p=0.026),  and LTO (p=0.002). Similarly, LCO has lower mean AUROC than LTO (p=0.015). In regression tasks the difference in PCC between LCO and LPO is significant (p=0.018). Due to the small sample sizes none of the other differences are statistically significant. Figure \ref{fig:perf}e shows the reported AUROC values per dataset. The distributions are relatively similar and no statistically significant differences can be shown between the datasets. 

\section{Discussion}

The majority of the synergy prediction methods focus on the LTO and LPO tasks, where the performance of the best methods is already very high, and probably hard to improve upon. Notably the top performers rely on large number of training examples combined with 'narrow' input representation, single input data types for drugs and cell lines.  The best results in the LCO and LDO tasks, on the other hand, are clearly lower, and suggest a shift of focus is needed for the method developers. In these scenarios, developing better representations for broad input data types could be a way forward.

We note that the evaluation setups across papers are not always clearly reported and easy to compare. Although some common protocols are in use, e.g. 5-fold cross-validation, the reporting of hyperparameter tuning is lacking in many papers, even among top performing models, which diminishes the confidence in the reported results. We are nevertheless happy to see the rigorous nested cross-validation approach in several papers.

Going forward, it seems clear that more unified approaches are needed for the inter-comparability of the methods. The standardization of benchmark datasets, prediction scenarios and evaluation protocols should help the community to make clearer assessment of the state-of-the-art and potential points of improvement.

\section{Acknowledgements}

The authors wish to acknowledge the financial support by Academy of Finland through the grants 339421 (MASF) and 345802 (AIB).

\end{document}